\def\year{2020}\relax
\documentclass[letterpaper]{article} 
\usepackage{aaai20}  
\usepackage{times}  
\usepackage{helvet} 
\usepackage{courier}  
\usepackage[hyphens]{url}  
\usepackage{graphicx} 
\usepackage{graphicx}  
\usepackage{epsfig}
\usepackage{amsmath}
\usepackage{amssymb}
\usepackage{textcomp}
\usepackage{multirow}
\usepackage[tight,footnotesize]{subfigure}
\usepackage{booktabs}
\usepackage{makecell}
\usepackage{xcolor}

\newcommand{\etal}{\textit{et al}.}

\title{
Adversarial Cross-Domain Action Recognition with Co-Attention
}

\author{Boxiao Pan\thanks{Equal contribution}, Zhangjie Cao\footnotemark[1], Ehsan Adeli, Juan Carlos Niebles \\
Stanford University \\
\texttt{\{bxpan, caozj, eadeli, jniebles\}@cs.stanford.edu}}
\begin{document}

\maketitle

\begin{abstract}
Action recognition has been a widely studied topic with a heavy focus on supervised learning involving sufficient labeled videos. However, the problem of cross-domain action recognition, where training and testing videos are drawn from different underlying distributions, remains largely under-explored. Previous methods directly employ techniques for cross-domain image recognition, which tend to suffer from the severe temporal misalignment problem. This paper proposes a Temporal Co-attention Network (TCoN), which matches the distributions of temporally aligned action features between source and target domains using a novel cross-domain co-attention mechanism. Experimental results on three cross-domain action recognition datasets demonstrate that TCoN improves both previous single-domain and cross-domain methods significantly under the cross-domain setting.
\end{abstract}

\section{Introduction}
Action recognition has long been studied in the computer vision community because of its wide range of applications in sports \cite{cite:Book14action}, healthcare \cite{cite:ICMLC18Health}, and surveillance systems \cite{cite:IJDSN16review}. Recently, motivated by the success of deep convolution networks in tasks on still images, such as image recognition \cite{cite:NIPS12AlexNet,cite:CVPR16ResNet} and object detection \cite{cite:CVPR15FastRCNN,cite:NIPS15FasterRCNN}, various deep architectures \cite{cite:ECCV16TSN,cite:ICCV15C3D} have been proposed for video action recognition. When large amounts of labeled videos are available, deep learning methods can achieve state-of-the-art performance on several benchmarks \cite{cite:DsetKinetics,cite:DsetHMDB,cite:DsetUCF}. 

\begin{figure}[t]
    \centering
    \includegraphics[width=.95\columnwidth]{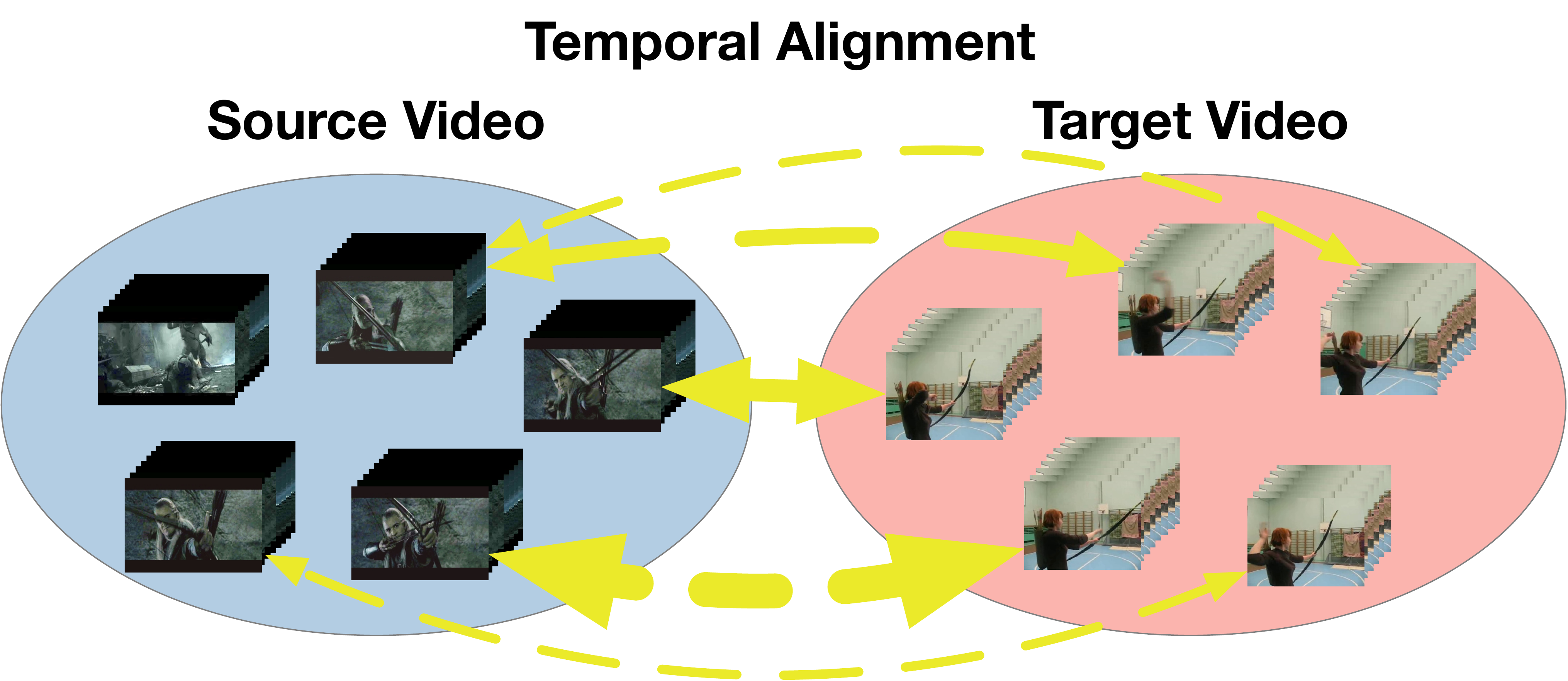}
    \caption{Illustration of our proposed temporal co-attention mechanism. Since different video segments represent distinct action stages, we propose to align segments that have similar semantic meanings. Key segment pairs that are more similar (denoted by thicker arrows) are assigned higher co-attention weights, thus contributing more during the cross-domain alignment stage. Here we show two samples with the action \textit{Archery} from HMDB51 (left) and UCF101 (right).}
    \label{fig:problem}
\end{figure}

Although current action recognition approaches achieve promising results, they mostly assume that the testing data follows the same distribution as the training data. Indeed, the performance of these models degenerate significantly when applied to datasets with different distributions due to domain shift. This greatly limits the application of current action recognition models. An example of domain shift is illustrated in Fig. \ref{fig:problem}, in which the source video is from a movie while the target video depicts a real-world scene. The challenge of domain shift motivates the problem of \textit{cross-domain} action recognition, where we have a target domain that consists of unlabeled videos and a source domain that consists of labeled videos. The source domain is related to the target domain but is drawn from a different distribution. Our goal is to leverage the source domain to boost the performance of action recognition models on the target domain.

The problem of cross-domain learning, also known as domain adaptation, has been explored on still image applications such as image recognition \cite{cite:ICML15DAN,cite:JMLR17DANN}, object detection \cite{cite:CVPR18DomainObjDec}, and semantic segmentation \cite{cite:ICML17CYCADA}. For still image problems, source and target domains differ mostly on appearance, and typical methods minimize a distribution distance within a latent feature space. However, for action recognition, source and target actions also differ temporally. For example, actions may appear at different time steps or last for different lengths in different domains. Thus, cross-domain action recognition requires matching action feature distributions between domains both spatially and temporally. Current action recognition methods typically generate features per-frame \cite{cite:ECCV16TSN} or per-segment \cite{cite:ICCV15C3D}. Previous cross-domain action recognition methods directly match segment feature distributions \cite{cite:BMVC18VideoDA} or with weights based on an attention mechanism \cite{chen2019temporal}. However, segment features can only represent parts of the action and may even be irrelevant to the action (e.g., background frames). Naively matching segment feature distributions would ignore the temporal order of segments and can introduce noisy matchings with background segments, which may be sub-optimal.

To address this challenge, we propose a Temporal Co-attention Network (TCoN). We first select segments that are critical for cross-domain action recognition by investigating temporal attention, which is a widely used technique in action recognition \cite{cite:Arxiv15Attention,cite:NIPS17Attentional} that helps the model focus on segments more related to the action. However, the vanilla attention mechanism fails when applied to the cross-domain setting. This is because many key segments are domain-specific, as they might not exist in one domain although they do in the other. Thus when calculating the attention score for a segment, apart from its self-importance, whether it matches with those in the other domain should also be taken into consideration. Only segments that are action-informative and also common in both domains should be paid close attention to. This motivates our design of a novel cross-domain co-attention module, which calculates attention scores for a segment based on both its action informativeness as well as cross-domain similarity. 

We further design a new matching approach by forming ``target-aligned source features" for target videos, which are derived from source features but aligned with target features temporally. Concatenating such target-aligned source segment features in temporal order naturally forms action features for source videos that are temporally aligned with target videos. Then, we match the distributions of the concatenated target-aligned source features with the concatenated target features to achieve cross-domain adaptation. Experimental results show that TCoN outperforms previous methods on several cross-domain action recognition datasets.

In summary, our main contributions are as follows: (1) We design a novel cross-domain co-attention module to concentrate the model on key segments shared by both domains, extending the traditional self-attention to cross-domain co-attention; (2) We propose a novel matching mechanism to enable distribution matching on temporally-aligned features; We also conduct experiments on three challenging benchmark datasets, and the results suggest that the proposed TCoN achieves state-of-the-art performance.

\section{Related Work}

\noindent\textbf{Video Action Recognition.}
With the success of deep Convolutional Neural Networks (CNNs) on image recognition \cite{cite:NIPS12AlexNet}, many deep architectures have been proposed to tackle action recognition from videos \cite{cite:CVPR16TwoStream,cite:ECCV16TSN,cite:ICCV15C3D,cite:CVPR17I3D,cite:CVPR18TRN,cite:Arxiv18TSM}. One branch of work is based on 2D CNNs. For instance, Two-Stream Network \cite{cite:CVPR16TwoStream} utilizes an additional optical flow stream to better leverage temporal information. Temporal Segment Network \cite{cite:ECCV16TSN} proposes a sparse sampling approach to remove redundant information. Temporal Relation Network \cite{cite:CVPR18TRN} further presents Temporal Relation Pooling to model frame relations at multiple temporal scales. Temporal Shifting Module \cite{cite:Arxiv18TSM} shifts feature channels along the temporal dimension to model temporal information efficiently. Another branch involves using 3D CNNs that learn spatio-temporal features. C3D \cite{cite:ICCV15C3D} directly extends the 2D convolution operation to 3D. I3D \cite{cite:CVPR17I3D} leverages pre-trained 2D CNNs such as ImageNet pre-trained Inception V1 \cite{cite:CVPR15Inception} by inflating 2D convolutional filters into 3D. However, all these works suffer from the spatial and temporal distribution gap between domains, which imposes challenges for cross-domain action recognition.

\noindent\textbf{Domain Adaption.}
Domain Adaptation aims to solve the cross-domain learning problem. In computer vision, previous work mostly focuses on still images. These methods fall into three categories. The first category focuses on minimizing distribution distances between source and target domains. DAN \cite{cite:ICML15DAN} minimizes the MMD distance between feature distributions. DANN \cite{cite:JMLR17DANN} and CDAN \cite{cite:NIPS18CDAN} minimize the Jensen-Shannon divergence between feature distributions with adversarial learning \cite{cite:NIPS14GAN}. The second category exploits techniques on semi-supervised learning. RTN \cite{cite:NIPS16RTN} exploits entropy minimization while Asym-Tri \cite{cite:ICML17PseudoLabel} uses pseudo labels. The third category groups all the image translation methods. Hoffman \etal~\cite{cite:ICML17CYCADA} and Murez \etal~\cite{cite:CVPR18IITrans} translate source labeled images to the target domain to enable supervised learning there. In this work, we adapt the first two categories of methods to videos as there is no sophisticated video translation method yet.

\noindent\textbf{Cross-Domain Action Recognition.} 
Despite the success of previous domain adaptation methods, cross-domain action recognition remains largely unexplored. Bian \etal~\cite{cite:System12CDAR} is the first to tackle this problem, which learns bag-of-words features to represent target videos and then regularizes the target topic model by aligning topic pairs across domains. However, their method requires partial target labeled data. Tang \etal~\cite{cite:Vision16CDAR} learns a projection matrix for each domain to map all features into a common latent space. Liu \etal~\cite{cite:TIP19MultiDTAC} employs more domains under the assumption that domains are bijective and trains the classifier with a multi-task loss. However, these methods assume deep video-level features available, which are not yet mature enough. Another work \cite{cite:BMVC18VideoDA} employs the popular GAN-based image domain adaptation approach to match segment features directly. Very recently, $\text{TA}^3\text{N}$ \cite{chen2019temporal} proposes to attentively adapt segments that contribute the most to the overall domain shift by leveraging the entropy of domain label predictor. However, both deep models suffer from temporal misalignment between domains since they only match segment features.

\begin{figure}[ht]
    \centering
    \includegraphics[width=.95\columnwidth]{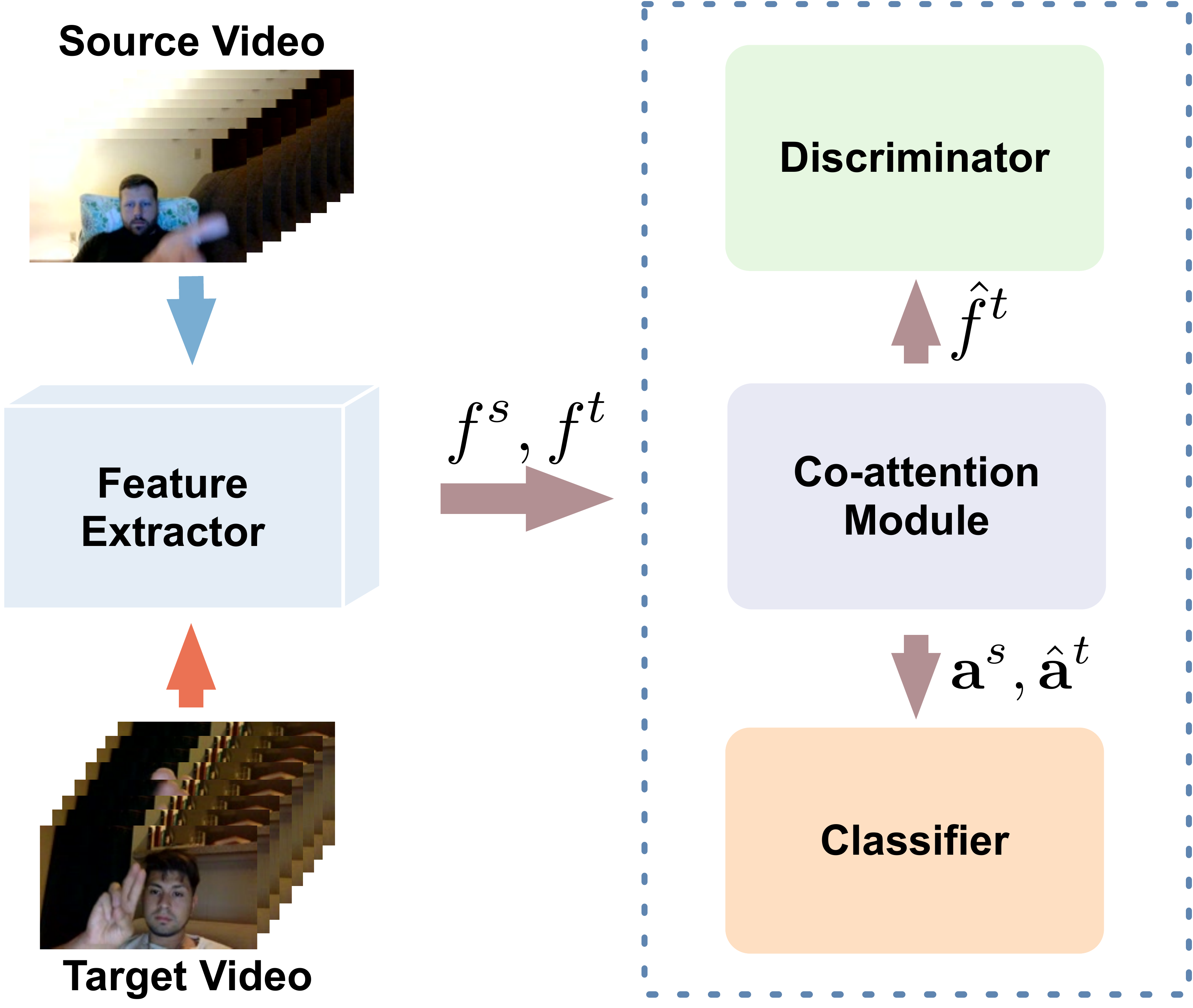}
    \caption{Our proposed TCoN framework. We first generate source and target features $f^s$ and $f^t$ with a feature extractor.
    The co-attention module uses $f^s$ and $f^t$ to generate ground-truth attention scores for source video ($\mathbf{a}^s$) and predicted score for target video ($\hat{\mathbf{a}}^t$), which the classifier uses to make predictions. At the same time, the co-attention module generates target-aligned source features $\hat{f}^t$, which the discriminator receives to enable temporally aligned distribution matching (see Figs. \ref{fig:co-attention-arch} and \ref{fig:discriminator-arch} for details).}
    \label{fig:arch}
\end{figure}

\section{Temporal Co-attention Network (TCoN)}
Suppose we have a source domain consisting of $N_s$ labeled videos $\mathcal{D}^s=\{V_i^s,y_i\}_{i=1}^{N_s}$ and a target domain consisting of $N_t$ unlabeled videos $\mathcal{D}^t=\{V_{i'}^t\}_{i'=1}^{N_t}$. The two domains are drawn from two different underlying distributions $p_s$ and $p_t$, but they are related and share the same label space. We will also provide analysis for the case where they do not share the same space in later section. The goal of cross-domain action recognition is to design an adaptation mechanism to transfer the recognition model learned from the source domain to the target domain with a low classification risk.

In addition to the appearance gap similar to the image case, there are two other main challenges that are specific to cross-domain action recognition. First, not all frames are useful under the cross-domain setting. Non-key frames contain noisy background information unrelated to the action, and even key frames can exhibit different cues for different domains. Second, current action recognition networks cannot generate holistic action features for the entire action in the video. Instead, they produce features of segments. Since segments are not temporally aligned between videos, it is hard to construct features for the entire action. To attack these two challenges, we design a co-attention module to focus on segments that contain important cues and are shared by both domains, which addresses the first challenge. We further leverage the co-attention module to generate temporally target-aligned source features. This enables distribution matching on temporally aligned features between domains, thus addressing the second challenge.

\subsection{Architecture}
Fig. \ref{fig:arch} shows the overall architecture of TCoN. During training, given a source and target video pair, we first partition each source video into $K_s$ segments and each target video into $K_t$ segments uniformly. We use $v_{ij}^s$ to denote the $j$-th segment in the $i$-th source video, and $v_{i'j'}^t$ is similarly defined for the target video. Then, we generate features $f_{ij}^{*}$ ($* \in (s,t)$) for each segment with a feature extractor $G_f$, which is a 2D or 3D CNN, i.e., $f_{ij}^{*}=G_f(v_{ij}^{*})$. Next, we use the co-attention module to calculate the co-attention matrix between this source and target video pair, from which we further derive the source and target attention score vectors, as well as the target-aligned source feature $\hat{f}^t$. Then, the discriminator module performs distribution matching given the source, target, and target-aligned source features. Finally, a shared classifier $G_y$ accepts the source and target features together with their attention scores, to predict the labels $\hat{y}_i^*$. For label prediction, we follow the standard practice, where we first predict per-segment labels and then weighted-sum segment predictions by attention scores.
\subsection{Cross-Domain Co-Attention}\label{sec:co-attention}
Co-attention was originally used in NLP to capture the interactions between questions and documents to boost the performance of question answering models \cite{cite:Arxiv16Coattention}. Motivated by this, we propose a novel cross-domain co-attention mechanism to capture correlations between videos from two domains. To the best of our knowledge, this is the first time that co-attention is explored under the cross-domain action recognition setting.

\begin{figure}[t]
    \centering
    \includegraphics[width=.95\columnwidth]{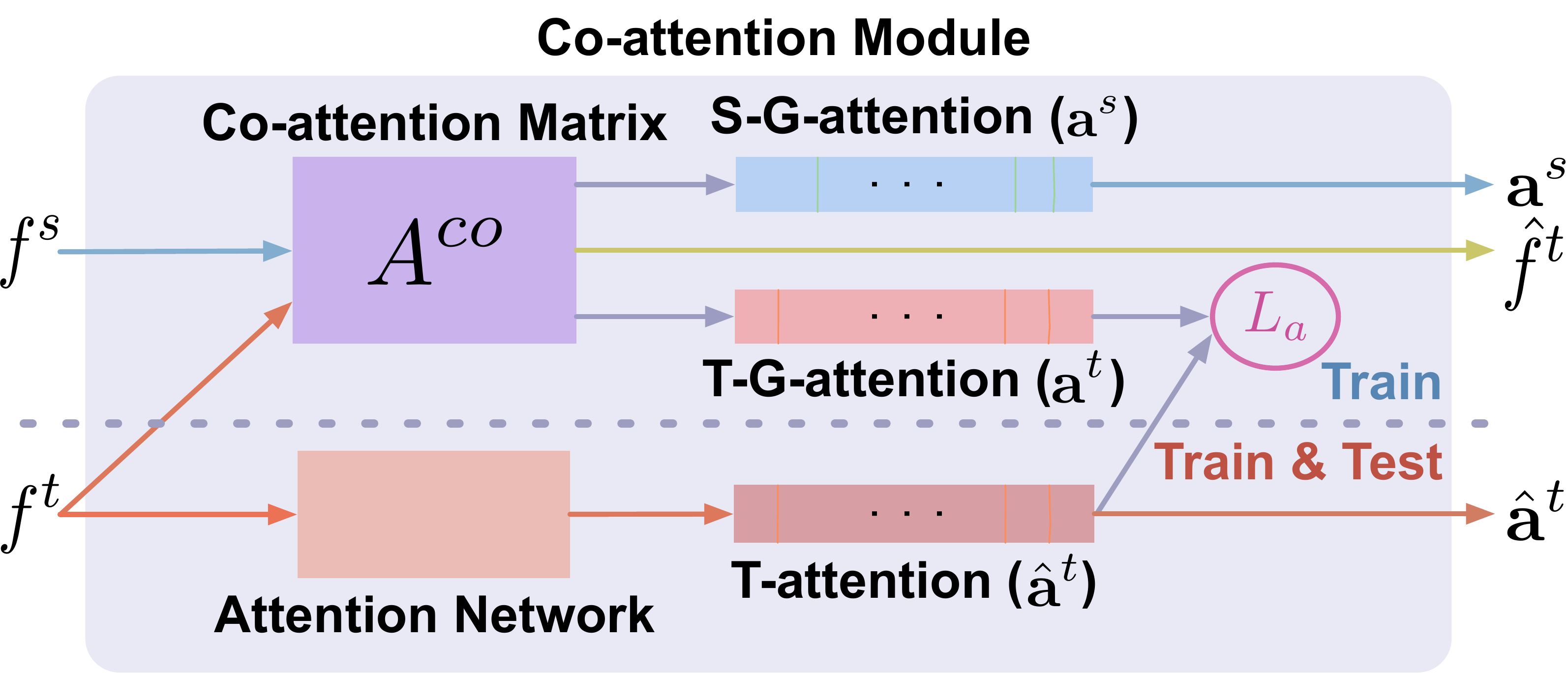}
    \caption{Structure of the co-attention module. It first calculates the co-attention matrix $A^{co}$ from source and target features $f^s$ and $f^t$. Then, source (S-G-attention) and target (T-G-attention) ground-truth attention $\mathbf{a}^s$ and $\mathbf{a}^t$ as well as target-aligned source features $\hat{f}^t$ are derived from $A^{co}$. $\mathbf{a}^t$ is used to train the target attention network, which predicts target attention (T-attention) $\hat{\mathbf{a}}^t$ for test time use.}
    \label{fig:co-attention-arch}
\end{figure}

The goal of the co-attention module is to model relations between source and target video pairs. As shown in Fig. \ref{fig:co-attention-arch}, given a pair of source and target videos $V_i^s$ and $V_{i'}^t$, and their segment features $f_{ij}^{s}|_{j=1}^{K_s}$ and $f_{i'j'}^{t}|_{j'=1}^{K_t}$, we use $k$ to index video pairs, i.e., $pair(k)=(i,i')$, which denotes that video pair $k$ is composed of the source video $V_i^s$ and the target video $V_{i'}^t$. We first calculate a self-attention score vector $\mathbf{a}_i^{ss}$ and $\mathbf{a}_{i'}^{tt}$ for each video:
\begin{equation} \label{eq:ss_attn}
    a_{ij}^{ss} = \frac{1}{K_s-1}\sum_{\bar{j}\ne j}\langle f_{i\bar{j}}^{s}, f_{ij}^{s} \rangle,
\end{equation}
\begin{equation} \label{eq:tt_attn}
    a_{i'j'}^{tt} = \frac{1}{K_t-1}\sum_{\bar{j}'\ne j'}\langle f_{i'\bar{j}'}^{t}, f_{i'j'}^{t} \rangle,
\end{equation}
where $a_{ij}^{ss}$ and $a_{i'j'}^{tt}$ are the $j$-th and $j'$-th element of $\mathbf{a}_i^{ss}$ and $\mathbf{a}_{i'}^{tt}$, respectively. $\langle \cdot, \cdot \rangle$ denotes the inner-product operation. Self-attention score vectors measure the intra-domain importance of a segment within a video. After we obtain these self-attention score vectors, we then derive each $(j, j')$-th element of the cross-domain similarity matrix $A_{k}^{st}$ as
    $a_{jj'}^{st} = \langle f_{ij}^{s}, f_{i'j'}^{t} \rangle$.
Note that for clarity, we drop the pair index $k$ for $a_{jj'}^{st}$. Each element $a_{jj'}^{st}$ measures the cross-domain similarity between segment pair $v_{ij}^s$ and $v_{i'j'}^t$. Finally, we calculate the cross-domain co-attention score matrix ${A}_{k}^{co}$ by
\begin{equation} \label{eq:st_attn}
    {A}_{k}^{co} = \left( \mathbf{a}_i^{ss}(\mathbf{a}_{i'}^{tt})^T \right) \odot A_{k}^{st},
\end{equation}
where $\odot$ represents element-wise multiplication. As can be seen from this process, in the co-attention matrix ${A}_{k}^{co}$, only the elements corresponding to those segment pairs $v_{ij}^s$ and $v_{i'j'}^t$ with high $a_{ij}^{ss}$, $a_{i'j'}^{tt}$ and $a_{jj'}^{st}$ would be assigned high values, which means that only key segment pairs that are also common in both domains will be paid high attention to. Thus, this co-attention matrix effectively reflects the correlations between source and target video pairs. Also, key segments (those with only high $a_{ij}^{ss}$ or $a_{i'j'}^{tt}$) or common segments (those with only high $a_{jj'}^{st}$) are not ignored, but paid less attention to. Only segments that are neither important nor common are discarded as they are essentially noise and do not help with the task.

For each segment, we derive attention scores from the above co-attention matrix by averaging its co-attention scores with all segments in the other video. We generate the ground-truth attention $\mathbf{a}_i^s$ and $\mathbf{a}_{i'}^t$ for $V_i^s$ and $V_{i'}^t$ as follows,
\begin{equation}
\small 
    \mathbf{a}_{i}^s = \frac{1}{M_i^s}\sum_{i \in pair(k)} \sum_{row} A_k^{co},
\end{equation}
\begin{equation}
    \mathbf{a}_{i'}^t = \frac{1}{M_{i'}^t}\sum_{i' \in pair(k)} \sum_{column} A_k^{co},
\end{equation}
where $\sum_{row}$ and $\sum_{column}$ are summation with respect to rows and columns, respectively. $M_i^s$ is the number of related pairs to $V_i^s$ and $M_{i'}^t$ is defined similarly. All the attention vectors sum to $1$. Since we should not assume access to source videos during testing time, we further use an attention network $G_a$, which is a fully-connected network, to predict attention scores for target videos:
{\small \begin{equation}
\hat{a}_{i'j'}^t = G_a(f_{i'j'}^{t}),
\end{equation}}
where $\hat{a}_{i'j'}^t$ is the $j'$-th element of the predicted attention. We further calculate the loss for the attention network with supervision from the ground-truth attention:
{\small \begin{equation}\label{eq:attention_loss}
    C_a = \frac{1}{N_t}\sum_{i'=1}^{N_t} L_a(\hat{\mathbf{a}}_{i'}^t, \mathbf{a}_{i'}^t),
\end{equation}
}where $L_a$ is the regression loss. With the source and target attention, the final classification loss is thus:

{\small \begin{equation} \label{eq:classification_loss}
\begin{aligned}
    C_y = &\frac{1}{N_s}\sum_{i}^{N_s}L_y(\sum_{j=1}^{K_s}a_{ij}^sG_y(f_{ij}^s), y_i^s) \\
    &+ \frac{1}{N_t}\sum_{i'}^{N_t}L_y(\sum_{j'=1}^{K_t}\hat{a}_{i'j'}^tG_y(f_{i'j'}^t), y_{i'}^t),
    \end{aligned}
\end{equation}}
where $L_y$ is the cross-entropy loss for classification. We train the classifier using source videos with ground-truth labels. Similar to previous work \cite{cite:ICML17PseudoLabel}, we also use target videos that have high label prediction confidence as training data, where the predicted pseudo-labels serve as the supervision. This helps preserve the $\lambda$ term in the error bound derived in Theorem 1 in \cite{cite:NIPS07DATheory}. Note that the total number of source and target video pairs is quadratic to the size of dataset, which is very large. For efficiency and co-attention with higher quality, we only calculate co-attention for video pairs with similar semantic information, i.e., video pairs with similar label prediction probability (which acts as a soft label).

\begin{figure}
\centering
    \centering
    \includegraphics[width=.95\columnwidth]{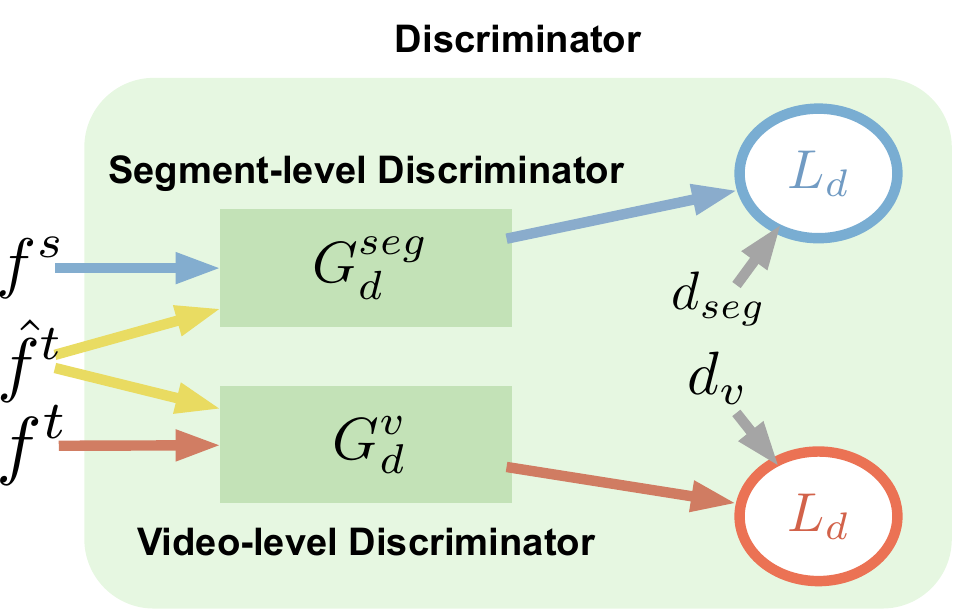}
    \caption{Structure of the discriminator. Source $f^s$ and target-aligned source $\hat{f}^t$ segment features are input to the segment-level discriminator to ensure that $\hat{f}^t$ does follow the source feature distribution. Target $f^t$ and target-aligned source $\hat{f}^t$ video features are input to the video-level discriminator to enable temporally aligned distribution matching.}
    \label{fig:discriminator-arch}
\end{figure}

\subsection{Temporal Adaptation}\label{sec:adaptation}
Fig. \ref{fig:discriminator-arch} illustrates the video-level discriminator we employ to match the distributions of target-aligned source and target video features, and the segment-level discriminator we use to match target-aligned source and source segment features.

For each pair $k$ of video $V_i^s$ and $V_{i'}^t$, the target-aligned source segment feature ${\hat{f}_{kj'}^t}|_{j'=1}^{K_t}$ is calculated as follows,
{\small \begin{equation}\label{eq:target-aligned}
    \hat{f}_{kj'}^t = \sum_{j=1}^{K_s}(A_{k}^{co})_{jj'} f_{ij}^s,
\end{equation}}
where $(A_{k}^{co})_{jj'}$ is the $(j, j')$-th element of $A_{k}^{co}$. Note that after we get $A_{k}^{co}$, we further normalize each column of it with a softmax function to keep the norm of target-aligned source features same as source features. Each target-aligned source segment feature is thus derived by weighted-summing source segment features, where the weight is the co-attention score between the target segment feature and each source segment feature. Thus, each target-aligned source segment feature preserves the semantic meaning of the corresponding target segment and falls on the source distribution. We concatenate segment features ${\hat{f}_{kj'}^t}|_{j'=1}^{K_t}$ as $\hat{F}_{k}^t$ and ${f_{i'j'}^t}|_{j'=1}^{K_t}$ as $F_{i'}^t$ in temporal order, which then naturally form as action features. Furthermore, $F_{i'}^t$ and all corresponding $\hat{F}_{k}^t$ are strictly temporally aligned since segments at the same time step express the same semantic meaning.

Then, we can derive the domain adversarial loss to match video distributions of target-aligned source features and target features. To further ensure the target-aligned source features to fall in the source feature space, we also match the segment distributions of target-aligned source features and source features. The loss for the video-level $C_{dv}$ and segment-level $C_{ds}$ discriminators are defined as follows,
{\small \begin{equation} \label{eq:d_ad_loss}
    C_{dv}=\frac{1}{N_t}\sum_{i'}L_d(G_d^v(F_{i'}^t), d_v) + \frac{1}{N_{st}}\sum_{k}L_d(G_d^v(\hat{F}_{k}^t), d_v),
\end{equation}
\begin{equation} \label{eq:s_adad_loss}
\begin{aligned}
    C_{ds} & =\frac{1}{N_sK_s}\sum_{i,j}L_d(G_d^{seg}(f_{ij}^s), d_{seg}) \\
    & + \frac{1}{N_{st}K_t}\sum_{k,j'}L_d(G_d^{seg}(\hat{f}_{kj'}^t), d_{seg}),
\end{aligned}
\end{equation}
}
where $N_{st}$ is the number of all pairs and $L_d$ is the binary cross-entropy loss. The video domain label $d_v$ is $0$ for target-aligned source features and $1$ for target features. The segment domain label $d_{seg}$ is $0$ for target-aligned source segment features and $1$ for target segment features.

\subsection{Optimization}
We perform optimization in the adversarial learning manner \cite{cite:JMLR17DANN}. We use $\theta_f$, $\theta_a$, $\theta_y$, $\theta_{d}^{v}$ and $\theta_{d}^{seg}$ to denote the parameters for $G_f$, $G_a$, $G_y$, $G_{d}^{v}$ and $G_{d}^{seg}$ and optimize:
{\small \begin{equation}
    \min_{\theta_f,\theta_a,\theta_y} C_y +\lambda_a C_a-\lambda_d(C_{dv}-C_{ds}),
\end{equation}
}
{\small
\begin{equation}
    \min_{\theta_{d}^{v},\theta_{d}^{seg}} C_{dv}+C_{ds},
\end{equation}
}
where $\lambda_a$ and $\lambda_d$ are trade-off hyper-parameters.

With the proposed Temporal Co-attention Network that contains an attentive classifier as well as a video- and a segment-level adversarial networks, we can simultaneously align the source and target video distributions while also minimize the classification error within both source and target domains, thus address the cross-domain action recognition problem in an effective way.

\section{Experiments}
Most prior work was done on small-scale datasets such as $\text{(UCF101-HMDB51)}_\text{1}$ \cite{cite:Vision16CDAR}, $\text{(UCF101-HMDB51)}_\text{2}$ \cite{chen2019temporal} and UCF50-Olympic\_Sports \cite{cite:BMVC18VideoDA}. For fair comparison with these works, we also evaluate our proposed method on these datasets. Moreover, we construct a large-scale cross-domain dataset, namely Jester (S)-Jester (T) (S for source, T for target), and further conduct experiments there. For $\text{(UCF101-HMDB51)}_\text{1}$, $\text{(UCF101- HMDB51)}_\text{2}$ and UCF50-Olympic\_Sports, we follow the prior works to construct the datasets by selecting the same action classes in two domains, while for Jester, we merge sub-actions into super-actions and split half of sub-actions into each domain. Please refer to the supplementary material for full details.

There are different types and extent of domain gap present in different datasets. For $\text{(UCF101-HMDB51)}_\text{1}$, $\text{(UCF101-HMDB51)}_\text{2}$ and UCF50-Olympic\_Sports, the domain gap is caused by appearance, lighting, camera viewpoint, etc., but not the action. Whereas for Jester, the gap arises from different action dynamics instead of other factors (since data samples from the same dataset but different sub-actions constitute a single super-action class). Hence, models trained on Jester suffer more from the temporal misalignment problem. Together with being at a larger scale, Jester is considered to be much harder than the other datasets.

We compare \textbf{TCoN} with single-domain methods (\textbf{TSN} \& \textbf{C3D} \& \textbf{TRN}) pre-trained on the source dataset, several cross-domain action recognition methods including a shallow learning method \textbf{CMFGLR} \cite{cite:Vision16CDAR}, deep learning methods \textbf{DAAA} \cite{cite:BMVC18VideoDA} and \textbf{$\text{TA}^3\text{N}$} \cite{chen2019temporal}, as well as a hybrid model which directly applies state-of-the-art domain adaptation method \textbf{CDAN} \cite{cite:NIPS18CDAN} to videos. We mainly use TSN \cite{cite:ECCV16TSN} as our backbone, but for a fair comparison with the prior work, we also conduct experiments using C3D \cite{cite:ICCV15C3D} and TRN \cite{cite:CVPR18TRN}.

\begin{table*}[t!]
\small
\centering
\caption{Accuracy (\%) of TCoN and compared methods on three datasets based on the TSN backbone. R + F denotes the results obtained by combining predictions from RGB and Flow models, which are calculated by averaging the logits (before softmax) from RGB and Flow models and then selecting the class with the highest entry in the obtained logits.}
\setlength{\tabcolsep}{5pt}
\begin{tabular}{l|ccc|ccc|ccc|ccc}
\Xhline{1pt}
\multirow{2}{40pt}{Method} & \multicolumn{3}{c|}{\scriptsize{$\text{(HMDB51} \rightarrow \text{UCF101)}_\text{1}$}}       & \multicolumn{3}{c|}{\scriptsize{UCF50 $\rightarrow$ Olympic\_Sports}} & \multicolumn{3}{c|}{\scriptsize{Olympic\_Sports $\rightarrow$ UCF50}} & \multicolumn{3}{c}{\scriptsize{Jester (S) $\rightarrow$ Jester (T)}} \\ \cline{2-13} 
& RGB & Flow & R + F & RGB & Flow & R + F & RGB & Flow & R + F & RGB & Flow & R + F \\
\hline
TSN \cite{cite:ECCV16TSN} & 82.10 & 76.86 & 83.11 & 80.00 & 81.82 & 81.75 & 76.67 & 73.34 & 74.47 & 51.70 & 49.89 & 50.56 \\
CMFGLR \cite{cite:Vision16CDAR} & 85.14 & 78.45 & 84.85 & 81.06 & 79.64 & 80.23 & 77.43 & 77.05 & 78.89 & 52.52 & 54.34 & 53.36 \\
DAAA  \cite{cite:BMVC18VideoDA} & 88.36 & 89.93 & 91.31 & 88.37 & 88.16 & 89.01 & 86.25 & 87.00 & 87.93 & 56.45 & 55.92 & 57.63 \\
CDAN \cite{cite:NIPS18CDAN} & 90.09 & 90.96 & 91.86 & 90.65 & 90.46 & 91.77 & 90.08 & 90.13 & 90.57 & 58.33 & 55.09 & 59.30 \\
TCoN (ours) & \textbf{93.01} & \textbf{96.07} & \textbf{96.78} & \textbf{93.91}  & \textbf{95.46} & \textbf{95.77} & \textbf{91.65}  & \textbf{93.77} & \textbf{94.12} & \textbf{61.78} & \textbf{71.11} & \textbf{72.24} \\
\Xhline{1pt}
\end{tabular}
\label{table:results_tsn}
\end{table*}

\begin{table*}[ht]
\small
\centering
\caption{Accuracy (\%) of TCoN and DAAA on two tasks with UCF50-Olympic\_Sports dataset based on C3D backbone.}
\begin{tabular}{l|ccc|ccc}
\Xhline{1pt}
\multirow{2}{40pt}{Method} & \multicolumn{3}{c|}{UCF50 $\rightarrow$ Olympic\_Sports} & \multicolumn{3}{c}{Olympic\_Sports $\rightarrow$ UCF50} \\ 
\cline{2-7} 
& RGB & Flow & R + F & RGB & Flow & R + F \\
\hline
C3D \cite{cite:ICCV15C3D} & 82.13 $\pm 0.52$ & 81.12 $\pm 0.85$ & 83.05 $\pm 0.65$ & 83.16 $\pm 0.75$ & 81.02 $\pm 0.97$ & 83.79 $\pm 0.82$ \\
DAAA \cite{cite:BMVC18VideoDA} & 91.60 $\pm 0.18$ & 89.16 $\pm 0.26$ & 91.37 $\pm 0.22$ & 89.96 $\pm 0.35$ & 89.11 $\pm 0.47$ & 90.32 $\pm 0.43$\\
TCoN (ours) & \textbf{94.73} $\pm 0.12$ & \textbf{96.03}$\pm 0.15$ & \textbf{95.92} $\pm 0.14$ & \textbf{92.88} $\pm 0.28$ & \textbf{94.25} $\pm 0.22$ & \textbf{94.77} $\pm 0.25$\\
\Xhline{1pt}
\end{tabular}
\label{table:results_c3d}
\end{table*}

\begin{table}[ht]
\small
\centering
\caption{Accuracy (\%) of TCoN and $\text{TA}^3\text{N}$ \cite{chen2019temporal} based on TRN backbone using only RGB input (U: UCF50 / UCF101, O: Olympic\_Sports, H: HMDB51).}
\setlength{\tabcolsep}{3pt}
\begin{tabular}{l|c|c|c|c|c}
\Xhline{1pt}
{Method} & {U $\rightarrow$ O} & {O $\rightarrow$ U} & {$\text{(U} \rightarrow \text{H)}_2$} & {$\text{(H} \rightarrow \text{U)}_2$} & {J(S) $\rightarrow$ J(T)} \\ 
\hline
$\text{TA}^3\text{N}$ & \textbf{98.15} & 92.92 & 78.33 & 81.79 & 60.11 \\
TCoN & 96.82 & \textbf{96.79} & \textbf{87.24} & \textbf{89.06} & \textbf{62.53} \\
\Xhline{1pt}
\end{tabular}
\label{table:results_ta3n}
\end{table}

\begin{table}[ht]
\small
\centering
\caption{Accuracy (\%) of TCoN compared with baselines when non-overlapping classes exist between domains.}
\begin{tabular}{l|c}
\Xhline{1pt}
{Method} & {$\text{(HMDB51} \rightarrow \text{UCF101)}_{\text{all}}$} \\ 
\hline
TSN \cite{cite:ECCV16TSN} & 66.81 \\
TRN \cite{cite:CVPR18TRN} & 68.07 \\
DAAA \cite{cite:BMVC18VideoDA} & 71.45 \\
TCoN (ours) & \textbf{75.23} \\
\Xhline{1pt}
\end{tabular}
\label{table:results_all}
\end{table}

\begin{table}[ht]
\small
\centering
\caption{Accuracy (\%) of TCoN and its variants.}
\begin{tabular}{l|ccc}
\Xhline{1pt}
\multirow{2}{*}{Method} & \multicolumn{3}{c}{Jester (S) $\rightarrow$ Jester (T)} \\ 
\cline{2-4} 
& RGB & Flow & R + F\\ 
\hline
TCoN - SAdNet & 61.23 & 68.23 & 71.13 \\
TCoN - TAdNet & 58.76 & 64.56 & 65.48 \\
TCoN - CoAttn & 57.25 & 56.93 & 57.95 \\
TCoN - Attn   & 59.03 & 62.74 & 63.13 \\
TCoN & \textbf{61.78} & \textbf{71.11} & \textbf{72.24} \\ 
\Xhline{1pt}
\end{tabular}
\label{table:ablation}
\end{table}

\subsection{Training Details}
For TSN, C3D and TRN, we train on source and test on target directly. For shallow learning methods, we use deep features from the source pre-trained model as input. For the hybrid model, we apply domain discriminator in CDAN to segment features and the consensus domain discriminator output of all segments as final output. For DAAA and $\text{TA}^3\text{N}$, we use their original training strategy. For TCoN, since the target attention network is not well-trained at the beginning, we use uniform attention at the first few iterations and plug it in when the loss of the attention network is lower than a certain threshold. To train TCoN more efficiently, we only calculate co-attention for segment pairs within mini-batchs.

We implement TCoN with the PyTorch framework \cite{cite:PyTorch}. We use the Adam optimizer \cite{kingma2014adam} and set the batch size to 64. For TSN and TRN-based models, we adopt the BN-Inception \cite{cite:BatchNorm} backbone pre-trained on ImageNet \cite{cite:ImageNet}. The learning rate is initialized to 0.0003 and decreases by $\frac{1}{10}$ every 30 epochs. We adopt the same data augmentation technique as in \cite{cite:ECCV16TSN}. For C3D-based models, we strictly follow the settings in \cite{cite:BMVC18VideoDA} and use the same base model \cite{cite:ICCV15C3D} pre-trained on Sports-1M dataset \cite{cite:sports1m}. We initialize the learning rate for the feature extractor to 0.001 while 0.01 for classifier since it is trained from scratch. The trade-off parameter $\lambda_d$ is increased gradually from $0$ to $1$ as in DANN \cite{cite:RevGrad}. For the number of segments, We do grid search for each dataset in \textit{[1, minimum video length]} on a validation set. Please refer to supplementary material for the actual numbers.

\begin{figure*}
\centering
\subfigure[DAAA segment features]{
  \centering
  \includegraphics[width=.48\columnwidth]{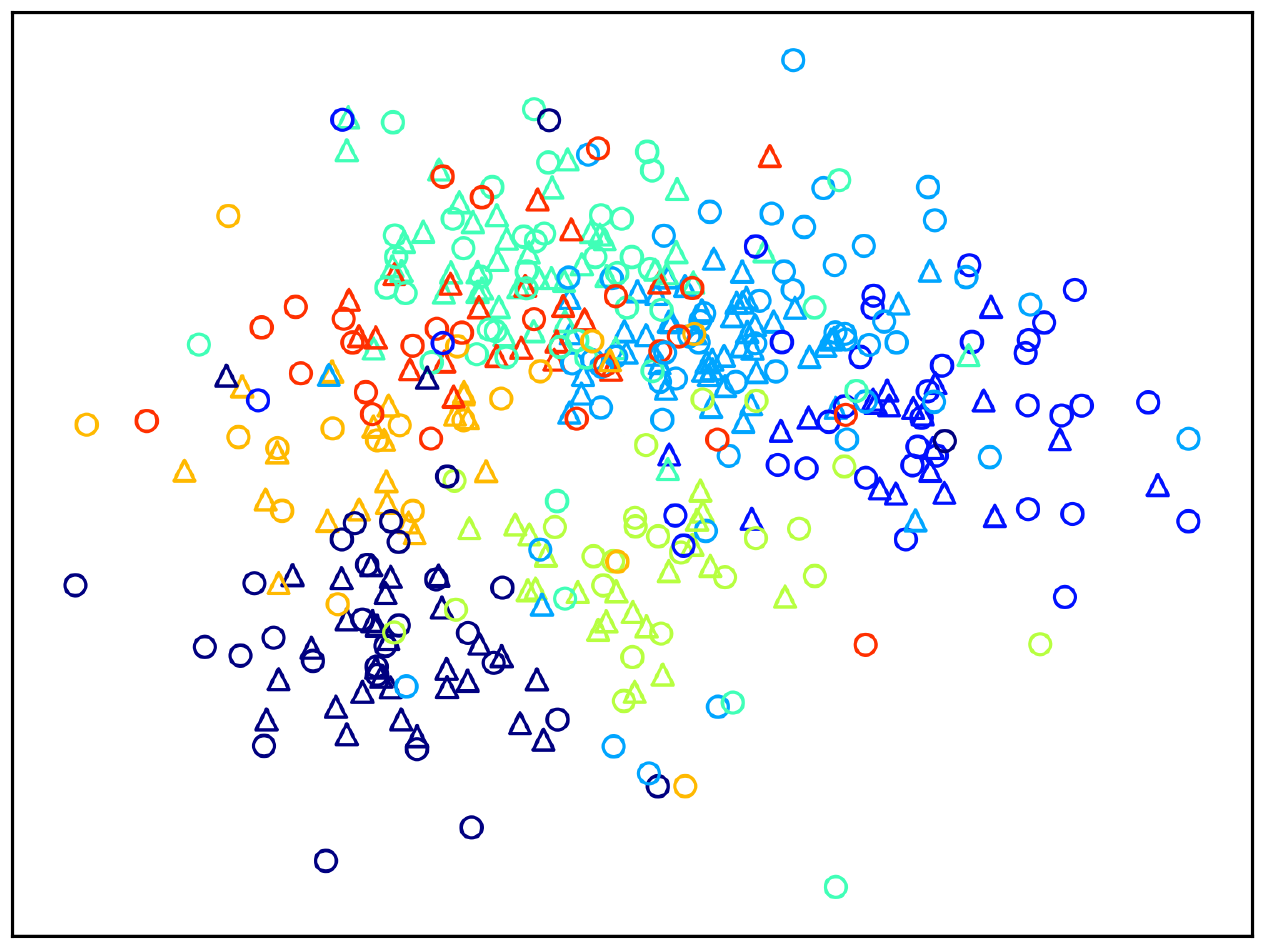}
  \label{fig:f_daaa}}
  \hfil
\subfigure[TCoN segment features]{
  \centering
  \includegraphics[width=.48\columnwidth]{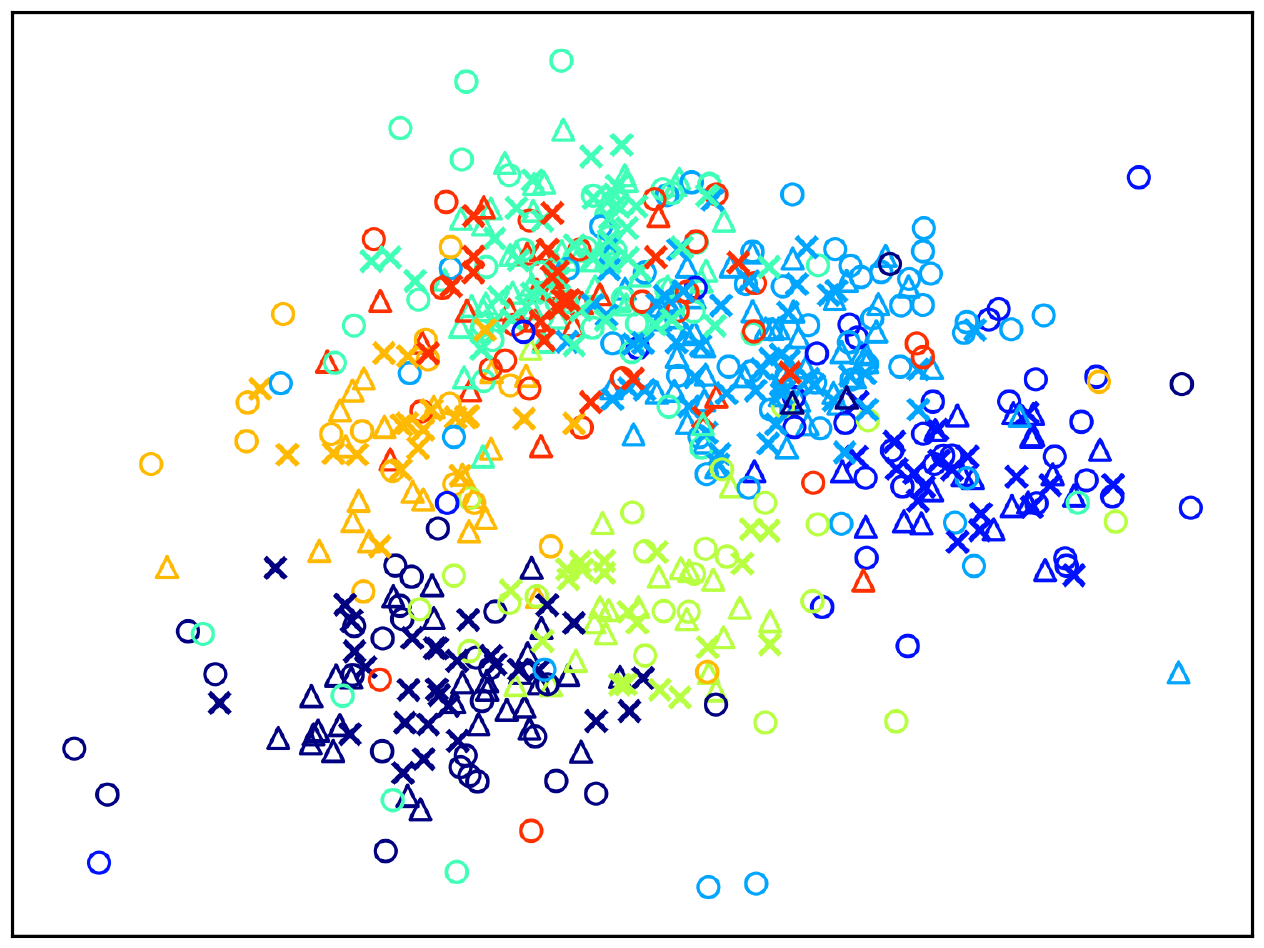}
  \label{fig:f_tcon}}
  \hfil
\subfigure[DAAA video features]{
  \centering
  \includegraphics[width=.48\columnwidth]{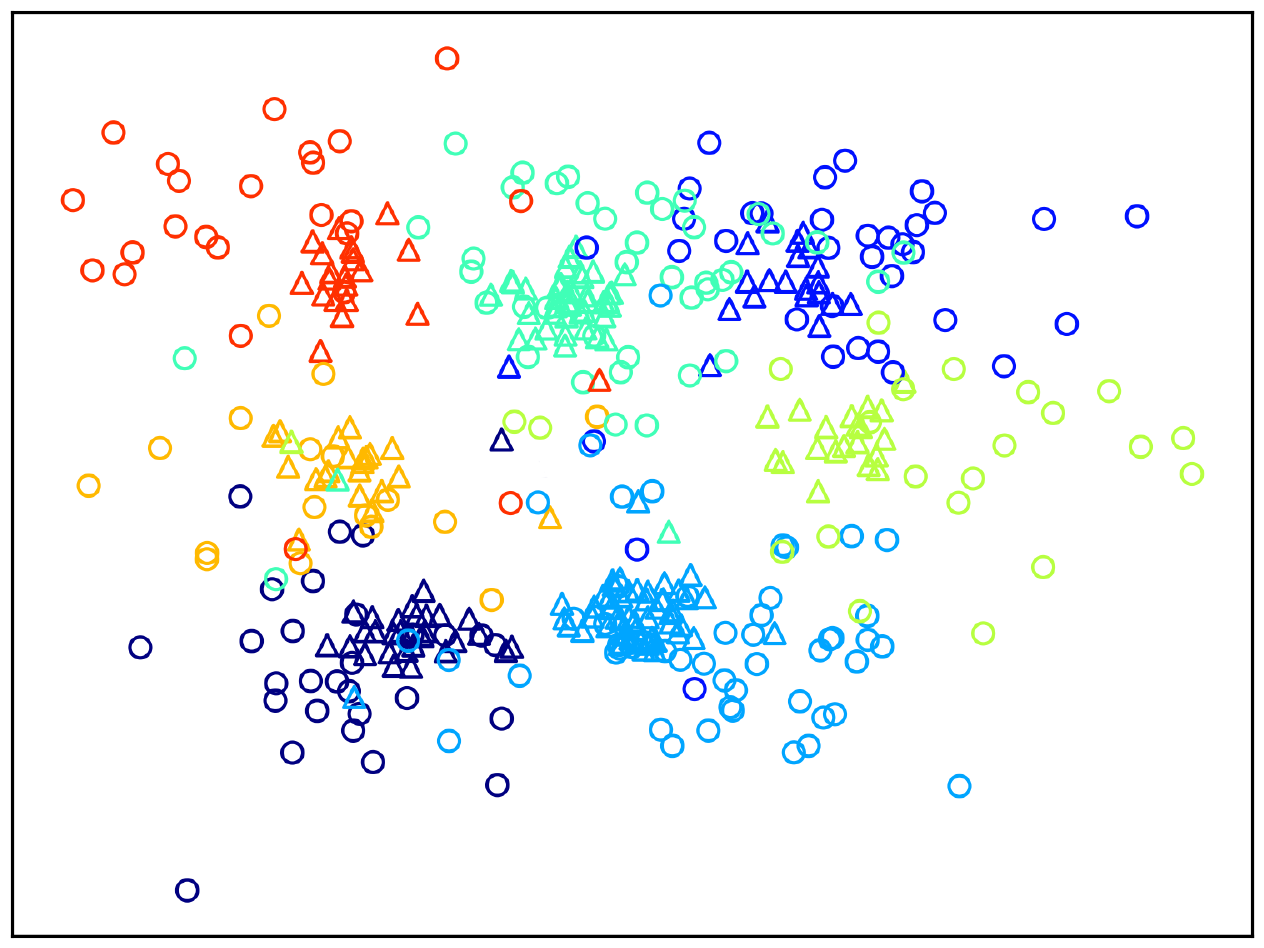}
  \label{fig:v_daaa}}
  \hfil
\subfigure[TCoN video features]{
  \centering
  \includegraphics[width=.48\columnwidth]{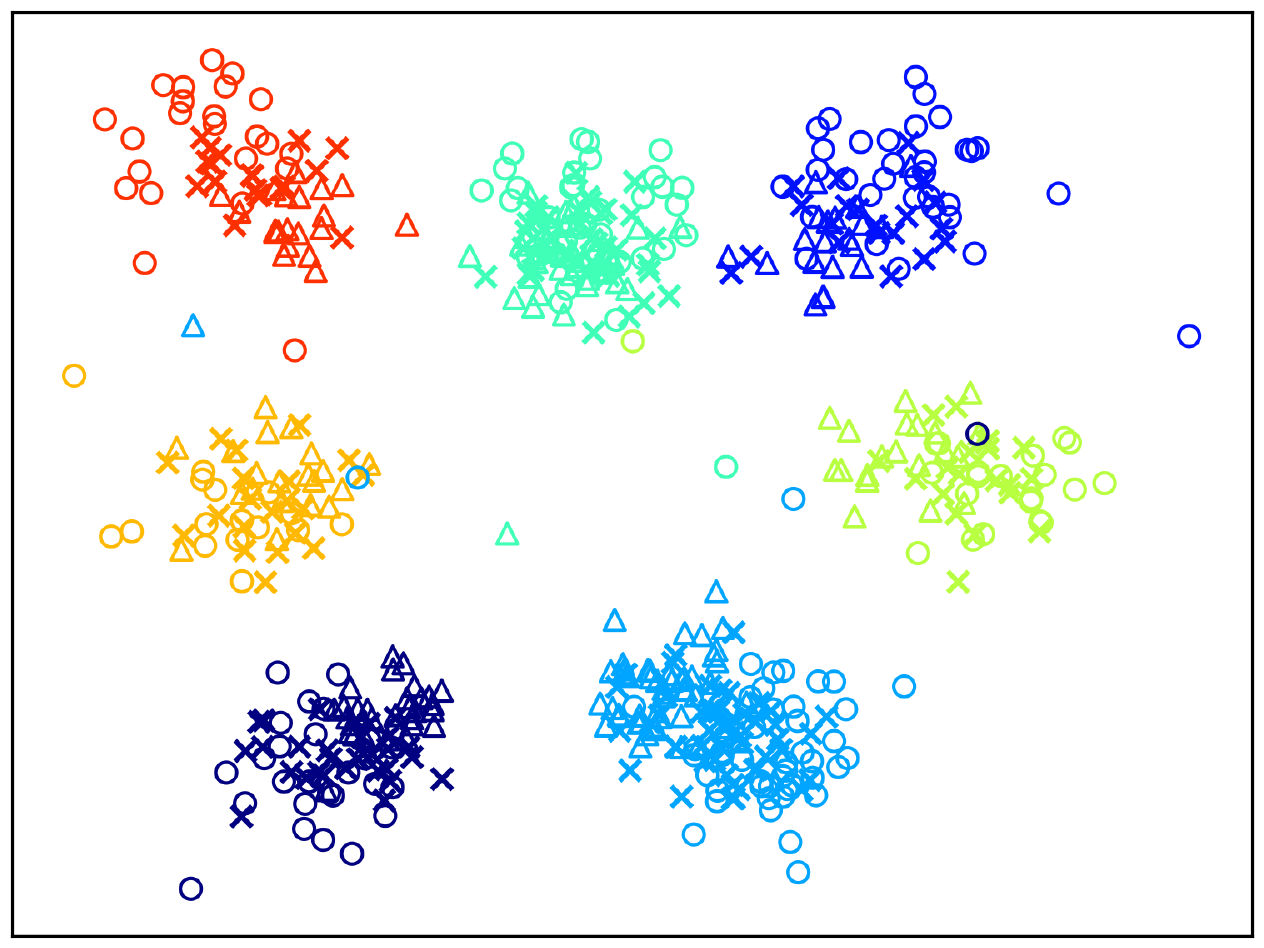}
  \label{fig:v_tcon}}
\caption{t-SNE visualization of features from DAAA \cite{cite:BMVC18VideoDA} and our TCoN. The first two figures are segment features, while the last two are video features. Different colors represent different classes. $\triangle$ stands for source features. $\circ$ denotes target features. $\times$ represents target-aligned source features.}
\label{fig:t-sne}
\end{figure*}

\subsection{Experimental Results}
The classification accuracies on three datasets using TSN-based TCoN are shown in Table \ref{table:results_tsn}. We can observe that the proposed TCoN outperforms all baselines on all datasets. In particular, TCoN improves previous methods with the largest margin on Jester, where temporal information is much more important, and temporal misalignment is more severe. Both CDAN and DAAA use segment features and minimize the Jensen-Shannon divergence of segment feature distributions between domains. The higher accuracy of TCoN demonstrates the importance of temporal alignment in distribution matching. We also notice that the Flow model consistently outperforms the RGB model for TCoN, indicating that TCoN well utilizes temporal information. 

We also compare with DAAA \cite{cite:BMVC18VideoDA} under their experiment setting with the C3D backbone. From Table \ref{table:results_c3d}, we can observe that TCoN outperforms DAAA on both tasks, which further suggests the efficacy of the proposed co-attention and distribution matching mechanism.

Moreover, we compare with the state-of-the-art cross-domain action recognition method $\text{TA}^3\text{N}$ \cite{chen2019temporal} on two datasets they used, namely UCF50-Olympic\_Sports and $\text{(UCF101-HMDB51)}_\text{2}$ (which is slightly different from $\text{(UCF101-HMDB51)}_\text{1}$ in 3 out of the 12 shared classes), as well as Jester. And we use the same backbone TRN \cite{cite:CVPR18TRN} and input modality (RGB) as theirs. The results from Table \ref{table:results_ta3n} show that on the datasets they used, TCoN outperforms $\text{TA}^3\text{N}$ on 3 out of 4 tasks and on par with it on the other. Moreover, TCoN achieves better performance on Jester, which again corroborates that TCoN can well handle not only the appearance gap but also the action gap.

To test whether our model is robust when the two domains do not share the same action space during training, we conduct experiments on $\text{(HMDB51} \rightarrow \text{UCF101)}_{\text{all}}$, where we train TCoN on data from all classes in HMDB51, but only test on the shared classes between two domains (it is impossible to predict those non-overlapping target classes). The results from Table \ref{table:results_all} show that TCoN still outperforms its baselines, suggesting the robustness of it in this case.

\begin{figure}
    \centering
    \includegraphics[width=.95\columnwidth]{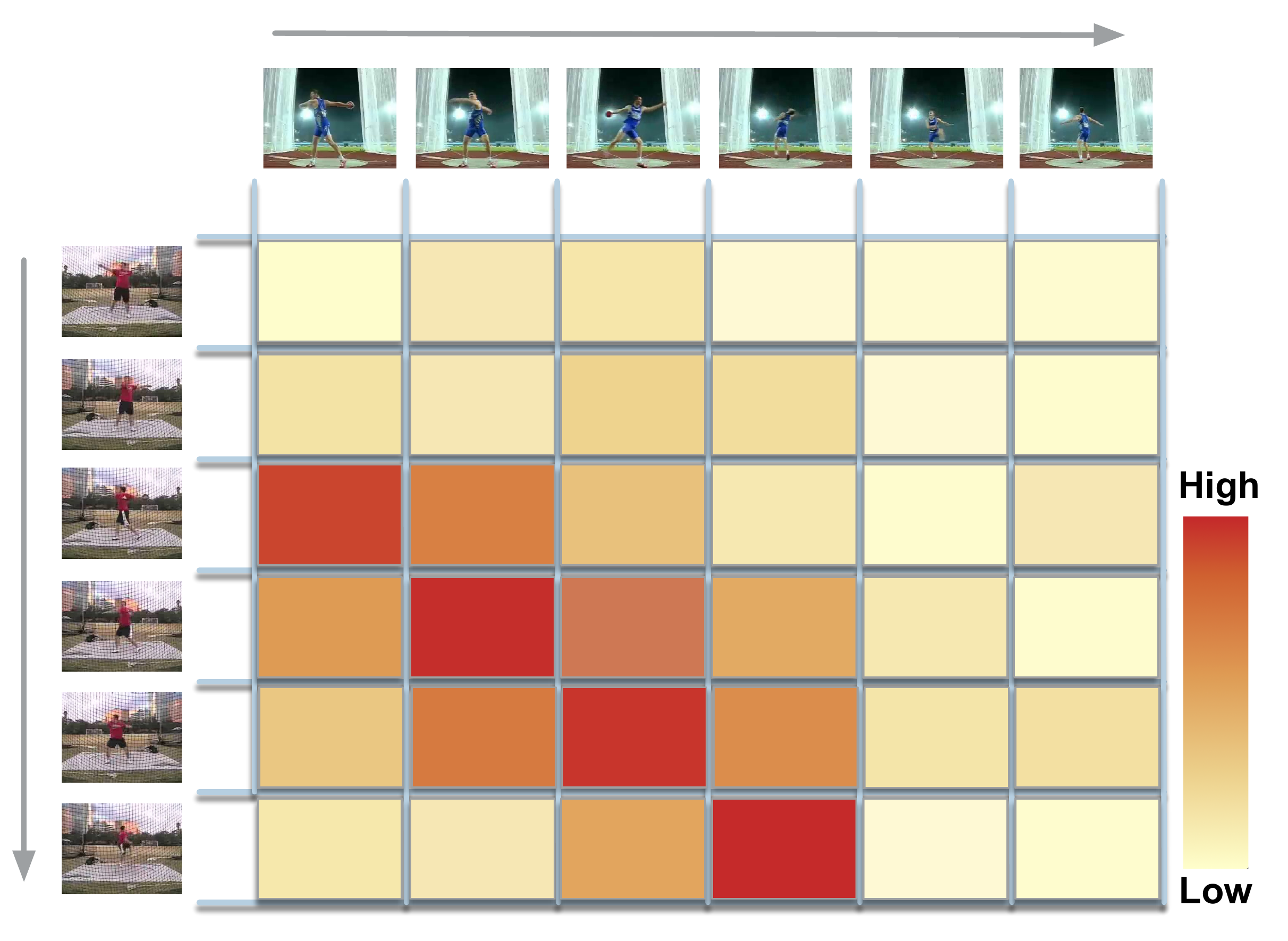}
    \caption{Co-attention matrix visualization for a video pair from UCF50 $\rightarrow$ Olympic\_Sports.}
    \label{fig:coattn_mat}
\end{figure}

\subsection{Analysis}
\noindent \textbf{Ablation Study.}
We compare TCoN with its four variants: (1) \textbf{TCoN - SAdNet} is the variant without the segment-level discriminator; (2) \textbf{TCoN - TAdNet} is the variant without using target-aligned source features but directly matching the source and concatenated target features with one discriminator; (3) \textbf{TCoN - CoAttn} is the variant with no co-attention computation. For the attentive classifier, we use self-attention instead; (4) \textbf{TCoN - AttnClassifier} is the variant where we directly average the classifier outputs from all segments instead of weighing them with attention scores generated from the co-attention matrix. The ablation study results on Jester (S) $\rightarrow$ Jester (T) are shown in Table \ref{table:ablation}, from which we can make the following observations: (1) TCoN outperforms TCoN-SAdNet on all modalities, demonstrating that the segment feature distribution for target-aligned source and source features are not exactly the same and a segment-level discriminator helps match the distributions; (2) TCoN outperforms TCoN-TAdNet by a large margin. This proves that target-aligned source features ease the temporal misalignment problem and improve distribution matching; (3) TCoN beats TCoN-CoAttn. This verifies the necessity of co-attention, which reflects both segment importance and similarity, in cross-domain action recognition; (4) TCoN outperforms TCoN-Attn, indicating that segments contribute to the prediction differently and it is crucial to focus on those informative ones.

\noindent \textbf{Visualization of Co-Attention.}
We further visualize the co-attention matrix between a video pair on the UCF50 $\rightarrow$ Olympic\_Sports dataset. The visualization is shown in Fig. \ref{fig:coattn_mat}, where the left video is from the source, and the top video is from the target. According to the co-attention matrix, we can observe that the first four frames in the target domain match the last four frames in the source domain, and the co-attention matrix assigns high values for these pairs. The first two frames of the source video show the person preparing his body for discus throw, which is not actually discus throw, thus are not considered as key frames. The last two frames of the target video are the ending stage of the action, which are important but do not exist in the source video. This proves that our co-attention mechanism exactly focuses attention on segments containing key action parts that are similar across source and target domains.

\noindent \textbf{Feature Visualization.}
We also plot the t-SNE embedding \cite{cite:ICML14decaf} for both segment and video features for DAAA and TCoN on Jester in Fig. \ref{fig:f_daaa} - \ref{fig:v_tcon}. For DAAA, we visualize the source (triangle) and target (circle) features. For TCoN, we visualize target-aligned source features (cross) as well. From Fig. \ref{fig:f_daaa} and \ref{fig:f_tcon}, we can observe that the segment features from different classes (shown in different colors) are mixed together, which is expected since segments cannot represent the entire action. In TCoN, the distributions of source and target-aligned source segment features are indistinguishable, demonstrating the effectiveness of our segment-level discriminator. From Fig. \ref{fig:v_daaa} and \ref{fig:v_tcon}, we can observe that for video features, TCoN has a better cluster structure than DAAA. In particular, in Fig. \ref{fig:v_tcon}, those points representing target-aligned source features lie between source and target feature points, suggesting that they actually bridge source and target features together. This sheds light on how the proposed distribution matching mechanism draws the target action distribution closer to the source by leveraging the target-aligned source features.

\section{Conclusion}
In this paper, we propose TCoN to address cross-domain action recognition. We design a cross-domain co-attention mechanism, which guides the model to pay more attention to common key frames across domains. We further introduce a temporally aligned distribution matching technique that enables distribution matching of action features. Extensive experiments on three benchmark datasets verify that our proposed TCoN achieves state-of-the-art performance.

\noindent\textbf{Acknowledgements.} The authors would like to thank Panasonic, Oppo, and Tencent for the support.

{\small
\bibliographystyle{aaai} 
\bibliography{cross_domain_action}

\begin{thebibliography}{}

\bibitem[\protect\citeauthoryear{Ben-David \bgroup et al\mbox.\egroup
  }{2007}]{cite:NIPS07DATheory}
Ben-David, S.; Blitzer, J.; Crammer, K.; and Pereira, F.
\newblock 2007.
\newblock Analysis of representations for domain adaptation.
\newblock In {\em NIPS}.

\bibitem[\protect\citeauthoryear{Bian, Tao, and Rui}{2012}]{cite:System12CDAR}
Bian, W.; Tao, D.; and Rui, Y.
\newblock 2012.
\newblock Cross-domain human action recognition.
\newblock {\em IEEE Transactions on Systems, Man, and Cybernetics, Part B
  (Cybernetics)}.

\bibitem[\protect\citeauthoryear{Carreira and Zisserman}{2017}]{cite:CVPR17I3D}
Carreira, J., and Zisserman, A.
\newblock 2017.
\newblock Quo vadis, action recognition? a new model and the kinetics dataset.
\newblock In {\em CVPR}.

\bibitem[\protect\citeauthoryear{Chen \bgroup et al\mbox.\egroup
  }{2018}]{cite:CVPR18DomainObjDec}
Chen, Y.; Li, W.; Sakaridis, C.; Dai, D.; and Van~Gool, L.
\newblock 2018.
\newblock Domain adaptive faster r-cnn for object detection in the wild.
\newblock In {\em CVPR}.

\bibitem[\protect\citeauthoryear{Chen \bgroup et al\mbox.\egroup
  }{2019}]{chen2019temporal}
Chen, M.-H.; Kira, Z.; AlRegib, G.; Woo, J.; Chen, R.; and Zheng, J.
\newblock 2019.
\newblock Temporal attentive alignment for large-scale video domain adaptation.
\newblock {\em arXiv preprint arXiv:1907.12743}.

\bibitem[\protect\citeauthoryear{Deng \bgroup et al\mbox.\egroup
  }{2009}]{cite:ImageNet}
Deng, J.; Dong, W.; Socher, R.; Li, L.-J.; Li, K.; and Fei-Fei, L.
\newblock 2009.
\newblock Imagenet: A large-scale hierarchical image database.

\bibitem[\protect\citeauthoryear{Donahue \bgroup et al\mbox.\egroup
  }{2014}]{cite:ICML14decaf}
Donahue, J.; Jia, Y.; Vinyals, O.; Hoffman, J.; Zhang, N.; Tzeng, E.; and
  Darrell, T.
\newblock 2014.
\newblock Decaf: A deep convolutional activation feature for generic visual
  recognition.
\newblock In {\em ICML}.

\bibitem[\protect\citeauthoryear{Feichtenhofer, Pinz, and
  Zisserman}{2016}]{cite:CVPR16TwoStream}
Feichtenhofer, C.; Pinz, A.; and Zisserman, A.
\newblock 2016.
\newblock Convolutional two-stream network fusion for video action recognition.
\newblock In {\em CVPR}.

\bibitem[\protect\citeauthoryear{Ganin and Lempitsky}{2014}]{cite:RevGrad}
Ganin, Y., and Lempitsky, V.
\newblock 2014.
\newblock Unsupervised domain adaptation by backpropagation.
\newblock {\em arXiv preprint arXiv:1409.7495}.

\bibitem[\protect\citeauthoryear{Ganin \bgroup et al\mbox.\egroup
  }{2017}]{cite:JMLR17DANN}
Ganin, Y.; Ustinova, E.; Ajakan, H.; Germain, P.; Larochelle, H.; Marchand, M.;
  and Lempitsky, V.
\newblock 2017.
\newblock Domain-adversarial training of neural networks.
\newblock {\em JMLR}.

\bibitem[\protect\citeauthoryear{Girdhar and
  Ramanan}{2017}]{cite:NIPS17Attentional}
Girdhar, R., and Ramanan, D.
\newblock 2017.
\newblock Attentional pooling for action recognition.
\newblock In {\em NIPS}.

\bibitem[\protect\citeauthoryear{Girshick}{2015}]{cite:CVPR15FastRCNN}
Girshick, R.
\newblock 2015.
\newblock Fast r-cnn.
\newblock In {\em ICCV}.

\bibitem[\protect\citeauthoryear{Goodfellow \bgroup et al\mbox.\egroup
  }{2014}]{cite:NIPS14GAN}
Goodfellow, I.; Pouget-Abadie, J.; Mirza, M.; Xu, B.; Warde-Farley, D.; Ozair,
  S.; Courville, A.; and Bengio, Y.
\newblock 2014.
\newblock Generative adversarial nets.
\newblock In {\em NIPS}.

\bibitem[\protect\citeauthoryear{He \bgroup et al\mbox.\egroup
  }{2016}]{cite:CVPR16ResNet}
He, K.; Zhang, X.; Ren, S.; and Sun, J.
\newblock 2016.
\newblock Deep residual learning for image recognition.
\newblock In {\em CVPR}.

\bibitem[\protect\citeauthoryear{Hoffman \bgroup et al\mbox.\egroup
  }{2017}]{cite:ICML17CYCADA}
Hoffman, J.; Tzeng, E.; Park, T.; Zhu, J.-Y.; Isola, P.; Saenko, K.; Efros,
  A.~A.; and Darrell, T.
\newblock 2017.
\newblock Cycada: Cycle-consistent adversarial domain adaptation.
\newblock {\em arXiv preprint arXiv:1711.03213}.

\bibitem[\protect\citeauthoryear{Ioffe and Szegedy}{2015}]{cite:BatchNorm}
Ioffe, S., and Szegedy, C.
\newblock 2015.
\newblock Batch normalization: Accelerating deep network training by reducing
  internal covariate shift.
\newblock {\em arXiv preprint arXiv:1502.03167}.

\bibitem[\protect\citeauthoryear{Jamal \bgroup et al\mbox.\egroup
  }{2018}]{cite:BMVC18VideoDA}
Jamal, A.; Namboodiri, V.~P.; Deodhare, D.; and Venkatesh, K.~S.
\newblock 2018.
\newblock Deep domain adaptation in action space.
\newblock In {\em BMVC}.

\bibitem[\protect\citeauthoryear{Karpathy \bgroup et al\mbox.\egroup
  }{2014}]{cite:sports1m}
Karpathy, A.; Toderici, G.; Shetty, S.; Leung, T.; Sukthankar, R.; and Fei-Fei,
  L.
\newblock 2014.
\newblock Large-scale video classification with convolutional neural networks.
\newblock In {\em CVPR}.

\bibitem[\protect\citeauthoryear{Kay \bgroup et al\mbox.\egroup
  }{2017}]{cite:DsetKinetics}
Kay, W.; Carreira, J.; Simonyan, K.; Zhang, B.; Hillier, C.; Vijayanarasimhan,
  S.; Viola, F.; Green, T.; Back, T.; Natsev, P.; et~al.
\newblock 2017.
\newblock The kinetics human action video dataset.
\newblock {\em arXiv preprint arXiv:1705.06950}.

\bibitem[\protect\citeauthoryear{Kingma and Ba}{2014}]{kingma2014adam}
Kingma, D.~P., and Ba, J.
\newblock 2014.
\newblock Adam: A method for stochastic optimization.
\newblock {\em arXiv preprint arXiv:1412.6980}.

\bibitem[\protect\citeauthoryear{Krizhevsky, Sutskever, and
  Hinton}{2012}]{cite:NIPS12AlexNet}
Krizhevsky, A.; Sutskever, I.; and Hinton, G.~E.
\newblock 2012.
\newblock Imagenet classification with deep convolutional neural networks.
\newblock In {\em NIPS}.

\bibitem[\protect\citeauthoryear{Kuehne \bgroup et al\mbox.\egroup
  }{2011}]{cite:DsetHMDB}
Kuehne, H.; Jhuang, H.; Garrote, E.; Poggio, T.; and Serre, T.
\newblock 2011.
\newblock Hmdb: a large video database for human motion recognition.
\newblock In {\em ICCV}.
\newblock IEEE.

\bibitem[\protect\citeauthoryear{Lin, Gan, and Han}{2018}]{cite:Arxiv18TSM}
Lin, J.; Gan, C.; and Han, S.
\newblock 2018.
\newblock Temporal shift module for efficient video understanding.
\newblock {\em arXiv preprint arXiv:1811.08383}.

\bibitem[\protect\citeauthoryear{Liu \bgroup et al\mbox.\egroup
  }{2019}]{cite:TIP19MultiDTAC}
Liu, A.-A.; Xu, N.; Nie, W.-Z.; Su, Y.-T.; and Zhang, Y.-D.
\newblock 2019.
\newblock Multi-domain and multi-task learning for human action recognition.
\newblock {\em IEEE Transactions on Image Processing}.

\bibitem[\protect\citeauthoryear{Long \bgroup et al\mbox.\egroup
  }{2015}]{cite:ICML15DAN}
Long, M.; Cao, Y.; Wang, J.; and Jordan, M.~I.
\newblock 2015.
\newblock Learning transferable features with deep adaptation networks.
\newblock In {\em ICML}.

\bibitem[\protect\citeauthoryear{Long \bgroup et al\mbox.\egroup
  }{2016}]{cite:NIPS16RTN}
Long, M.; Zhu, H.; Wang, J.; and Jordan, M.~I.
\newblock 2016.
\newblock Unsupervised domain adaptation with residual transfer networks.
\newblock In {\em NIPS}.

\bibitem[\protect\citeauthoryear{Long \bgroup et al\mbox.\egroup
  }{2018}]{cite:NIPS18CDAN}
Long, M.; Cao, Z.; Wang, J.; and Jordan, M.~I.
\newblock 2018.
\newblock Conditional adversarial domain adaptation.
\newblock In {\em NIPS}.

\bibitem[\protect\citeauthoryear{Murez \bgroup et al\mbox.\egroup
  }{2018}]{cite:CVPR18IITrans}
Murez, Z.; Kolouri, S.; Kriegman, D.; Ramamoorthi, R.; and Kim, K.
\newblock 2018.
\newblock Image to image translation for domain adaptation.
\newblock In {\em CVPR}.

\bibitem[\protect\citeauthoryear{Ogbuabor and La}{2018}]{cite:ICMLC18Health}
Ogbuabor, G., and La, R.
\newblock 2018.
\newblock Human activity recognition for healthcare using smartphones.
\newblock In {\em ICMLC}.
\newblock ACM.

\bibitem[\protect\citeauthoryear{Paszke \bgroup et al\mbox.\egroup
  }{2017}]{cite:PyTorch}
Paszke, A.; Gross, S.; Chintala, S.; Chanan, G.; Yang, E.; DeVito, Z.; Lin, Z.;
  Desmaison, A.; Antiga, L.; and Lerer, A.
\newblock 2017.
\newblock Automatic differentiation in pytorch.

\bibitem[\protect\citeauthoryear{Ranasinghe, Al~Machot, and
  Mayr}{2016}]{cite:IJDSN16review}
Ranasinghe, S.; Al~Machot, F.; and Mayr, H.~C.
\newblock 2016.
\newblock A review on applications of activity recognition systems with regard
  to performance and evaluation.
\newblock {\em IJDSN}.

\bibitem[\protect\citeauthoryear{Ren \bgroup et al\mbox.\egroup
  }{2015}]{cite:NIPS15FasterRCNN}
Ren, S.; He, K.; Girshick, R.; and Sun, J.
\newblock 2015.
\newblock Faster r-cnn: Towards real-time object detection with region proposal
  networks.
\newblock In {\em NIPS}.

\bibitem[\protect\citeauthoryear{Saito, Ushiku, and
  Harada}{2017}]{cite:ICML17PseudoLabel}
Saito, K.; Ushiku, Y.; and Harada, T.
\newblock 2017.
\newblock Asymmetric tri-training for unsupervised domain adaptation.
\newblock In {\em ICML}.
\newblock JMLR. org.

\bibitem[\protect\citeauthoryear{Sharma, Kiros, and
  Salakhutdinov}{2015}]{cite:Arxiv15Attention}
Sharma, S.; Kiros, R.; and Salakhutdinov, R.
\newblock 2015.
\newblock Action recognition using visual attention.
\newblock {\em arXiv preprint arXiv:1511.04119}.

\bibitem[\protect\citeauthoryear{Soomro and Zamir}{2014}]{cite:Book14action}
Soomro, K., and Zamir, A.~R.
\newblock 2014.
\newblock Action recognition in realistic sports videos.
\newblock In {\em Computer vision in sports}. Springer.

\bibitem[\protect\citeauthoryear{Soomro, Zamir, and Shah}{2012}]{cite:DsetUCF}
Soomro, K.; Zamir, A.~R.; and Shah, M.
\newblock 2012.
\newblock Ucf101: A dataset of 101 human actions classes from videos in the
  wild.
\newblock {\em arXiv preprint arXiv:1212.0402}.

\bibitem[\protect\citeauthoryear{Szegedy \bgroup et al\mbox.\egroup
  }{2015}]{cite:CVPR15Inception}
Szegedy, C.; Liu, W.; Jia, Y.; Sermanet, P.; Reed, S.; Anguelov, D.; Erhan, D.;
  Vanhoucke, V.; and Rabinovich, A.
\newblock 2015.
\newblock Going deeper with convolutions.
\newblock In {\em CVPR}.

\bibitem[\protect\citeauthoryear{Tang \bgroup et al\mbox.\egroup
  }{2016}]{cite:Vision16CDAR}
Tang, J.; Jin, H.; Tan, S.; and Liang, D.
\newblock 2016.
\newblock Cross-domain action recognition via collective matrix factorization
  with graph laplacian regularization.
\newblock {\em Image Vision Comput.} (P2).

\bibitem[\protect\citeauthoryear{Tran \bgroup et al\mbox.\egroup
  }{2015}]{cite:ICCV15C3D}
Tran, D.; Bourdev, L.; Fergus, R.; Torresani, L.; and Paluri, M.
\newblock 2015.
\newblock Learning spatiotemporal features with 3d convolutional networks.
\newblock In {\em ICCV}.

\bibitem[\protect\citeauthoryear{Wang \bgroup et al\mbox.\egroup
  }{2016}]{cite:ECCV16TSN}
Wang, L.; Xiong, Y.; Wang, Z.; Qiao, Y.; Lin, D.; Tang, X.; and Van~Gool, L.
\newblock 2016.
\newblock Temporal segment networks: Towards good practices for deep action
  recognition.
\newblock In {\em ECCV}.
\newblock Springer.

\bibitem[\protect\citeauthoryear{Xiong, Zhong, and
  Socher}{2016}]{cite:Arxiv16Coattention}
Xiong, C.; Zhong, V.; and Socher, R.
\newblock 2016.
\newblock Dynamic coattention networks for question answering.
\newblock {\em arXiv preprint arXiv:1611.01604}.

\bibitem[\protect\citeauthoryear{Zhou \bgroup et al\mbox.\egroup
  }{2018}]{cite:CVPR18TRN}
Zhou, B.; Andonian, A.; Oliva, A.; and Torralba, A.
\newblock 2018.
\newblock Temporal relational reasoning in videos.
\newblock In {\em ECCV}.

\end{thebibliography}
}

\end{document}